\pdfoutput=1
\documentclass[10pt,logo,copyright]{nvidiatechreport}
\usepackage[authoryear,sort&compress,round]{natbib}
\setcitestyle{numbers}
\usepackage{algorithm}
\usepackage{algorithmic}
\usepackage{bm}
\usepackage{xspace}
\usepackage{tikz}
\usepackage{subcaption}
\usepackage{multirow}
\usepackage[capitalise,nameinlink]{cleveref}

\crefname{figure}{Fig.}{Figs.}
\Crefname{figure}{Fig.}{Figs.}
\hypersetup{
  pdftitle={ACCRUE: Continual Off-Policy Reinforcement Learning for Robot Foundation Models},
  pdfauthor={Changyeon Kim, Kimin Lee, Yashraj Narang, Yijie Guo}
}

\newcommand{\eqcomment}[1]{\COMMENT{\labelcref{#1}}}   
\newcommand{\refa}{\tilde{\mathbf{a}}}
\newcommand{\base}{\pi_{\text{base}}} 
\newcommand{\method}{\mbox{\fontfamily{JetBrainsMono}\bfseries\upshape ACCRUE}\xspace}
\newcommand{\metfull}{\textit{\underline{A}ctor–\underline{C}ritic \underline{C}arryover for continual off-policy \underline{R}einforcement learning \underline{U}sing retained \underline{E}xperience}\xspace}
\title{\method: Continual Off-Policy Reinforcement Learning for Robot Foundation Models}
\author[1,2]{Changyeon Kim}
\author[2]{Kimin Lee}
\author[1]{Yashraj Narang}
\author[1]{Yijie Guo}
\affil[1]{NVIDIA}
\affil[2]{KAIST}
\makeatletter
\g@addto@macro\AB@affillist{\\[6pt]%
  \makebox[\linewidth][l]{\makebox[\textwidth][c]{\normalfont\fontfamily{lmtt}\fontsize{11}{13}\selectfont
    \href{https://accrue-rl.github.io/}{https://accrue-rl.github.io/}}}}
\makeatother

\begin{abstract}
Reinforcement learning (RL) post-training improves robot foundation models (RFMs) beyond their demonstrations, but real-world deployment demands more: learning new tasks efficiently while retaining previously acquired behaviors. Existing continual RL methods on top of VLAs mitigate forgetting but rely on on-policy updates with limited forward transfer, whereas off-policy methods improve single-task sample efficiency without addressing learner persistence across tasks. We propose \method, a continual off-policy RL framework that carries lightweight actor and critic networks across tasks, allowing each stage to reuse behavioral and value knowledge acquired through prior RL. The inherited critic adapts only on current-task transitions and guides current-task actor learning, while prior behavior is preserved by distilling the actor toward cached targets on a small number of retained episodes. Fixed visual features keep the inherited learner and its replay compatible with base-policy updates. Across contact-rich assembly sequences in simulation and on real-robot hardware, \method accelerates subsequent-task learning with near-zero forgetting and significantly reduces real-robot training time by 31.0\% to 68.5\%.
\end{abstract}

\begin{document}
\maketitle
\abscontent

\section{Introduction}\label{sec:intro}

\begin{figure}[!t]
    \centering
    \includegraphics[width=0.65\linewidth]{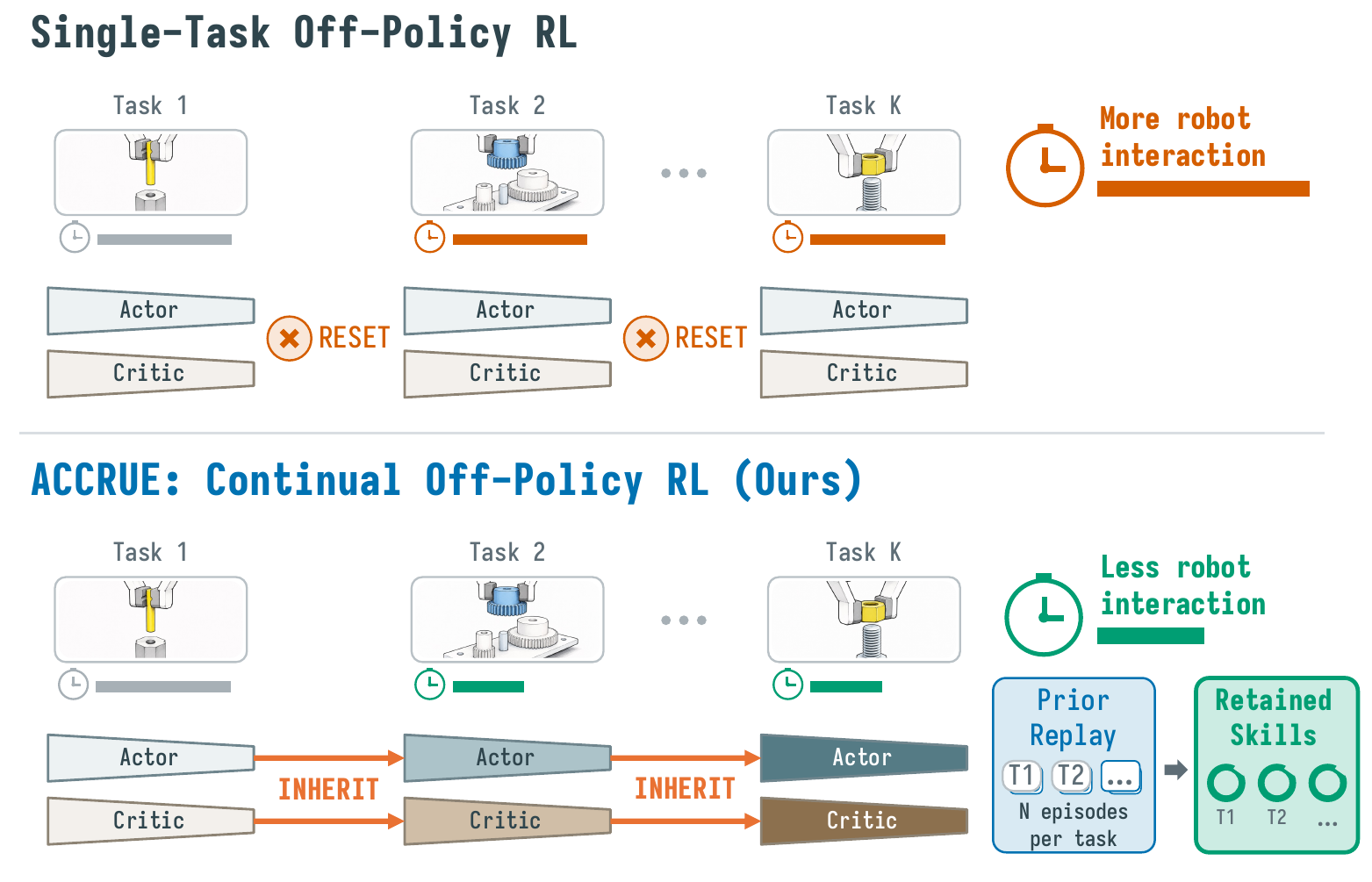}

    \caption{\textbf{Single-task vs Continual off-policy RL.} Single-task RL resets the actor and critic for each task, requiring repeated online interactions. Our method inherits both networks and uses small prior-task replay to accelerate learning while retaining prior performance.}
    \label{fig:setting}
\end{figure}

Robot foundation models (RFMs), including vision-language-action models (VLAs)~\cite{zitkovich2023rt,kim2024openvla,black2024pi0,nvidia2025groot} and world action models (WAMs)~\cite{ye2026dreamzero,dyna2026dyna2,zhu2025uwm,pai2025mimicvideo}, have demonstrated strong performance across a wide range of manipulation tasks. Despite these generalization capabilities, RFMs remain imperfect at deployment, where robots encounter tasks and conditions not covered by pretraining. Reinforcement learning (RL) post-training can close this gap by improving a model beyond its demonstrations through online interaction~\cite{li2025gr,xiao2026self,li2026simplevla,xu2026rlt,dong2026expoft}. Most prior work, however, treats post-training as a one-time procedure for a single target task. A robot deployed in the real world must instead learn to perform new tasks throughout its operation, often from only a few demonstrations and without access to future tasks in advance. Robotic assembly illustrates this need: new parts and product variants require robots to adapt their contact behaviors while retaining skills for previously encountered assemblies. Restarting RL for every task repeatedly pays the full interaction cost and fails to exploit knowledge acquired during earlier RL stages.
Carrying the learner across tasks introduces the risk of forgetting previously acquired behaviors.
Practical continual post-training must therefore \textit{reduce} the interaction required for each new task without sacrificing capabilities acquired from previous tasks (\Cref{fig:setting}).

Recent methods use RL to adapt VLAs across sequential tasks and mitigate forgetting~\cite{hu2026vlacrl,liu2026towards,zeng2026crl}, but provide limited forward transfer. Whereas on-policy updates require fresh current-policy rollouts, off-policy learning can reuse previously collected transitions to improve interaction efficiency. Existing approaches such as RL Token~\cite{xu2026rlt} and EXPO-FT~\cite{dong2026expoft}, which attach lightweight actor and critic networks to frozen RFMs, improve single-task sample efficiency but do not address learner persistence across tasks.

Directly extending these approaches across tasks introduces practical challenges. Retaining the full interaction history becomes increasingly costly as tasks accumulate, motivating a small replay memory for preserving prior behavior. With limited prior-task coverage, adapting an inherited critic to a new task can make its estimates unreliable on earlier tasks; using those estimates to drive actor updates on prior-task samples can then erode previously learned behavior. Meanwhile, updating the base RFM can change its internal representations and action proposals, potentially shifting the inputs consumed by the inherited networks.

To this end, we propose \method (\metfull), a continual off-policy RL framework that inherits a lightweight actor and critic across tasks to amortize prior RL experience. The inherited actor and critic transfer behavioral and value priors, respectively, enabling subsequent tasks to be learned with fewer environment interactions. Starting from a pretrained RFM, we fine-tune the base policy at each stage on demonstrations observed so far and hold it fixed during RL.
At each stage, the inherited critic is adapted only to the current-task transitions. We deliberately exclude prior-task samples from critic updates because the small number of retained episodes may not provide sufficient coverage for reliable value estimation on all prior tasks.
The adapted critic guides actor improvement on the current task but is not used to update the actor on prior tasks. Instead, \method preserves prior behavior by distilling the actor toward cached outputs of the preceding-stage actor on the last \(N\) episodes retained from each completed task (20--100 episodes per task in main experiments). This asymmetric update retains the benefits of actor–critic inheritance without relying on the critic to preserve prior-task behavior. A fixed external visual encoder provides consistent visual features across base-policy updates, while reference-action conditioning allows the lightweight learner to refine proposals from the current base policy.

We showcase \method on contact-rich assembly sequences in simulation and on real-robot hardware, with experiments covering VLA and WAM base policies and different off-policy RL algorithms. In simulation, \method accelerates subsequent-task learning while limiting forgetting of prior tasks. Across FORGE and AutoMate, it improves continual AUC by up to 21.0\% and reduces negative backward transfer by up to 70.0\% compared with the strongest off-policy continual baseline. Ablations and baseline comparisons show that critic inheritance supports forward transfer and that separating current-task RL from prior-task actor preservation improves retention under limited replay. On two multi-stage real-robot sequences, \method reduces robot RL interaction time for subsequent tasks by 31.0\%–68.5\% compared with training a new actor–critic for each task, while maintaining approximately 97\% average success on prior tasks.
\section{Related Work}

\subsection{RL Post-Training for RFMs}

RL post-training improves RFMs beyond demonstration-only supervision using task rewards and deployment experience. One line of work directly updates the VLA or its action head through policy-gradient learning, typically requiring fresh current-policy rollouts and substantial parallel interaction~\cite{guo2025improving,li2026simplevla}. Other approaches use offline value learning, task-specific residual policies, or RL-generated data to improve a generalist policy~\cite{kim2026deas,li2025gr,xiao2026self}. Most closely related to our architecture, lightweight off-policy methods refine a frozen policy through action- or latent-space actor--critic learning~\cite{yuan2025policy,wagenmaker2025dsrl,xu2026rlt,dong2026expoft}, but these methods focus on improving sample efficiency only on individual tasks. Our work considers the complementary problem of carrying a lightweight off-policy learner across sequentially arriving tasks while the underlying RFM is repeatedly updated with new demonstrations.

\subsection{Continual Learning for Robotics}

Continual robot learning~\cite{liu2023libero,wan2024lotus} has largely focused on preserving demonstrated behavior through experience replay~\cite{liu2026forgetmenot,du2026memory}, hierarchical skill libraries~\cite{wan2024lotus}, task-specific adapters~\cite{liu2024tail}, or parameter merging~\cite{yadav2026robust}. Beyond imitation learning, recent continual VLA methods carry policy parameters across tasks using on-policy RL~\cite{hu2026vlacrl,zeng2026crl} or demonstration-derived reinforcement fine-tuning~\cite{liu2026towards}. The former require fresh interaction at each stage, whereas the latter do not transfer value knowledge acquired through environment interaction. Concurrent work CIDER learns a shared visuomotor policy from task-specific data without a pretrained RFM, combining current-task off-policy human-in-the-loop RL with prior-task actor distillation~\cite{li2026cider}. In contrast, \method builds on manipulation skills already learned by a pretrained VLA or WAM and uses RL to improve its proposed actions. By retaining the actor and critic across tasks, \method reuses earlier RL experience even as the base model is further fine-tuned with new demonstrations.
\begin{figure}[!t]
    \centering
    \includegraphics[width=\textwidth]{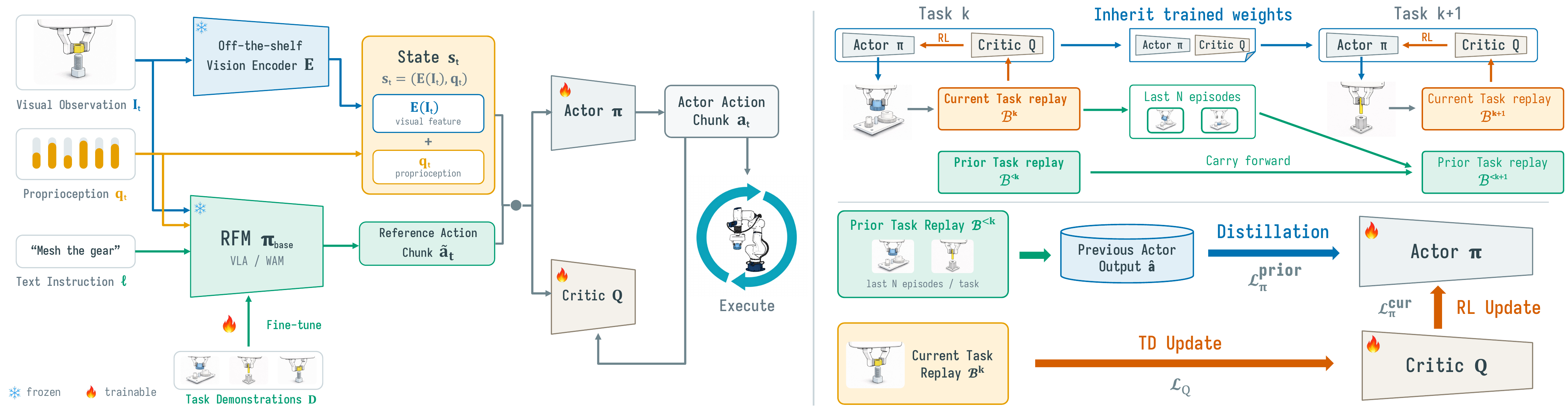}

    \caption{\textbf{Overview of \method.} (Left) Base policy is fine-tuned on demonstrations observed up to current stage, held fixed, and proposes reference action chunks that actor and critic networks refine through off-policy RL on frozen encoder features. (Right Top) Both networks are carried across stages through a handoff, independent of the base policy, with small fraction of online replay buffer. (Right Bottom) While updating the current task with RL objectives, earlier tasks are preserved by distilling the actor toward cached targets on a small retained memory.}
    \label{fig:overview}
\end{figure}

\section{Preliminaries}
\label{sec:prelim}

\subsection{Robot Foundation Models}
\label{subsec:rfm}
We build on a base policy $\base$, a robot foundation model (RFM) pretrained on large-scale robot data, such as VLA~\cite{zitkovich2023rt,kim2024openvla,black2024pi0,nvidia2025groot} or WAM~\cite{ye2026dreamzero,dyna2026dyna2,zhu2025uwm,pai2025mimicvideo}. Given multi-view images $\mathbf{I}_t$, proprioception $\mathbf{q}_t$, and a language instruction $\ell$, the policy outputs an $H$-step action chunk $\tilde{\mathbf{a}}_t = (\tilde{a}_t, \dots, \tilde{a}_{t+H-1}) \in \mathbb{R}^{H \times d}$, where $d$ denotes the dimension of the action space~\cite{zhao2023act,chi2025diffusion}. We use the same horizon $H$ for prediction and execution in all experiments. We access $\base$ only through this input--output interface, so our framework is agnostic to the choice of internal architecture in $\base$.

\subsection{Problem Setup}
\label{subsec:problem}

We consider a robot that acquires $K$ tasks sequentially, one task per stage. Each task $T^{k}$ is a finite-horizon Markov decision process that shares its state space, action space, and discount $\gamma$ with the other tasks, but may differ in its transition function, reward function, initial-state distribution, and horizon. Reward function $r_t$ is sparse and emitted upon task success. We follow a task-incremental continual-learning setting~\cite{van2022three}, where task boundaries and identities are known. 

At stage $k$, a small expert demonstration set $\mathcal{D}^{k}$ (1--50 trajectories in our experiments) becomes available for $T^{k}$. The base policy is fine-tuned on all demonstrations observed up to that stage, $\mathcal{D}^{1:k}=\bigcup_{i=1}^{k}\mathcal{D}^{i}$, yielding $\base^{k}$. Starting from $\base^{k}$, the agent performs RL using new interactions collected exclusively from the current task $T^{k}$ and stored in the current-task replay buffer $\mathcal{B}^{k}$. The learner additionally uses a bounded prior-task replay buffer $\mathcal{B}^{<k}$, which contains a subset of online experience retained from earlier RL stages. Training on $\mathcal{B}^{k}$ and $\mathcal{B}^{<k}$ produces the adapted policy $\pi^{k}$. Unlike protocols~\cite{hu2026vlacrl, liu2026towards} that fine-tune the base policy on the complete task suite before continual RL begins, our setup assumes neither access to future demonstrations nor a fixed base policy throughout the sequence. It therefore captures repeated post-training as new task requirements and demonstrations arrive over time.

After stage $k$, $\pi^{k}$ is evaluated on all tasks encountered so far, with the corresponding language instruction provided for each task. The objective is to maximize the average return across these tasks:
\begin{equation*}
\max_{\pi^{k}} \; J^{k}(\pi^{k}) =
\frac{1}{k}\sum_{i=1}^{k}
\mathbb{E}_{\tau \sim (\pi^{k},T^{i})}
\left[\sum_{t=0}^{L_{\tau}}\gamma^{t}r^{i}_{t}\right],
\label{eq:objective}
\end{equation*}
where $L^{\tau}$ denotes the length of trajectory $\tau$. Although new training interactions can be collected only from $T^{k}$, the learner should exploit knowledge acquired during earlier RL stages to adapt efficiently to the current task while preserving performance on previous tasks.

\subsection{Action-Chunk Reinforcement Learning}
\label{subsec:chunk_rl}
Within stage $k$, we improve performance on the current task $T^{k}$ using an off-policy actor--critic algorithm~\cite{seo2025fasttd3,seo2025learning} trained on the current-task replay buffer $\mathcal{B}^{k}$.
Let $s_t$ denote the state available at timestep $t$. The actor $\pi_\theta(\cdot \mid s_t)$ outputs an $H$-step action chunk $\mathbf{a}_t=(a_t,\dots,a_{t+H-1})$, which is executed in full before the next policy query. The critic $Q_\psi(s_t,\mathbf{a}_t)$ takes the state and the full action chunk as inputs to estimate its value~\cite{li2026reinforcement,kim2026deas,xu2026rlt,dong2026expoft}. We define $R_t^{(H)} = \sum_{h=0}^{H-1} \gamma^{h} r_{t+h}$ as the $H$-step discounted reward sum collected during chunk execution. The chunk value is estimated by the $H$-step backup
\begin{equation*}
  Q_\psi(s_t, \mathbf{a}_t) \approx R_t^{(H)} + \gamma^{H}\,\mathbb{E}_{\mathbf{a}' \sim \pi_\theta(\cdot \mid s_{t+H})}\!\left[Q_\psi(s_{t+H}, \mathbf{a}')\right].
  \label{eq:chunk_backup}
\end{equation*}
Relative to single-action backups, chunk-level backups reduce the nominal credit-assignment path by a factor of $H$.

\section{Method}
\label{sec:method}

We introduce \method, a simple but effective continual off-policy RL framework for RFMs (Fig.~\ref{fig:overview}). As new task demonstrations arrive at each stage, the base policy is fine-tuned on demonstrations from all tasks encountered so far and then held fixed throughout the RL phase of that stage.
The lightweight actor and critic are conditioned on visual features from a fixed external encoder and reference action chunks proposed by the current base policy (Sec.~\ref{subsec:interfaces}).
The fixed encoder keeps the visual representation unchanged across base-policy updates. For stage $k>1$, the actor and critic are initialized from the preceding checkpoint (Sec.~\ref{subsec:current}). With limited replay, the critic may no longer accurately evaluate actions on earlier tasks. We therefore train the critic only on current-task transitions and preserve prior behavior by matching the current actor's outputs to those of the preceding-stage actor on retained samples (Sec.~\ref{subsec:continual}).

\subsection{Actor--Critic Inputs}
\label{subsec:interfaces}

To carry the actor and critic across stages, their visual inputs must keep a consistent representation as the base policy is updated.
We therefore use an off-the-shelf visual encoder $E$ that is independent of the base policy and remains frozen across stages.
We form the actor--critic input from the encoder output $E(\mathbf{I}_t)$ and robot proprioception $\mathbf{q}_t$ as $s_t = \big(E(\mathbf{I}_t),\, \mathbf{q}_t\big)$.

The actor and the critic apply their own lightweight trainable projector to $s_t$. The projectors are parameterized as part of $\theta$ and $\psi$, respectively. The replay buffer $B^k$ stores $s_t$ rather than the outputs from these trainable projectors.
Freezing $E$ keeps the visual features of every stored transition in the same representation across stages, allowing the inherited projectors to process inputs $s_t$ collected at earlier stages. 

The input $s_t$ does not explicitly encode the language instruction. 
The known task identity determines the language instruction $\ell^k$ provided to the base policy, which produces the reference action chunk $\refa_t$.
Rather than encoding the instruction again in the actor and critic, we condition both networks on the instruction-conditioned reference to obtain $\mathbf{a}_t \sim \pi_\theta(s_t, \refa_t)$ and $Q_\psi\big(s_t, \mathbf{a_t}, \refa_t\big)$.
  

Because the base policy already encodes task semantics and provides a task-level action proposal, the lightweight actor learns a local refinement conditioned on $\refa$ rather than reconstructing the full skill. Continual learning therefore needs to retain only the RL-acquired refinement, reducing the amount of behavior that must remain stable in the shared actor. Moreover, related tasks can share refinement and value structure, allowing the inherited actor and critic to provide useful initializations for subsequent tasks. Removing reference conditioning increases forgetting at a similar final success rate (Table~\ref{tab:ablation}). Unlike RL Token~\cite{xu2026rlt}, which conditions only the actor on $\refa$ , we condition the critic as well,
because the action value is task-dependent and the reference provides task contexts.
To keep the actor from merely copying the reference, we apply reference action dropout~\cite{xu2026rlt} during training, replacing $\refa$ with zeros for a random fraction $p$ of each batch for both networks.

\subsection{Current-Task Off-Policy RL}
\label{subsec:current}

Actor and critic network parameters are randomly initialized only at the first stage. For each stage $k>1$, we initialize the actor--critic pair from the checkpoint of stage $k-1$, transferring a behavioral initialization through the actor and a value initialization through the critic. 
The current-task buffer $\mathcal{B}^k$ is seeded with warmup episodes collected by executing the base policy $\base^k$. The actor then interacts with the environment, and the actor--critic pair is updated on transitions sampled from $\mathcal{B}^k$.

The actor objective is to maximize the value estimated by the critic and remain close to the reference chunk,
\begin{equation}
  \mathcal{L}_{\pi}^{\mathrm{cur}}(\theta ; \mathcal{B}^{k}) = \mathbb{E}_{\substack{
  (s,\refa) \sim \mathcal{B}^k \\[-2pt]
  {}\mathbf{a} \sim \pi_{\theta}(s, \refa)
  }}
  \big[
    -Q_\psi\big(s,\mathbf{a},\refa\big)
    + \beta \lVert \mathbf{a}-\refa \rVert_2^2
  \big],
  \label{eq:actor_current}
\end{equation}
where $\beta$ penalizes deviation from the reference action~\cite{xu2026rlt}, reducing the learning to local refinement of the base policy's action distribution rather than exhaustive search over $Hd$-dimensional action spaces.

The critic is trained with off-policy temporal-difference learning.
For a transition $(s_t,\allowbreak \mathbf{a}_t,\allowbreak \refa_t,\allowbreak R_t^{(H)},\allowbreak s_{t+H},\allowbreak \refa_{t+H})$ sampled from $\mathcal{B}^k$, we write the common chunk-level target for value bootstrapping, omitting learner-specific terms:
\begin{equation}
  \hat{Q}
  =
  R_t^{(H)}+\gamma^H Q_{\psi'}(s_{t+H},\mathbf{a}_{t+H},\refa_{t+H}),
\label{eq:td_loss}
\end{equation}
where $\psi'$ denotes the target-critic parameters. The next action chunk $\mathbf{a}_{t+H}$ is sampled from the actor for stochastic learners or formed using the target-action rule for deterministic learners.
In implementation, the critic predicts a categorical return distribution over a fixed support, and $Q_\psi$ denotes its expected value. We train the critic with cross-entropy loss~\cite{bellemare2017distributional}, which we found more stable under sparse reward. 

\subsection{Continual Operation}
\label{subsec:continual}

At the end of stage $k-1$, \method transfers the actor--critic parameters $(\theta^{k-1}, \psi^{k-1})$ and $\mathcal{B}^{<k}$, a memory of retained experience, to stage $k$. The base policy is separately fine-tuned on $\mathcal{D}^{1:k}$ to obtain $\base^k$ and then held fixed during stage-$k$ RL. Fine-tuning on accumulated demonstrations is intended to retain the base policy's earlier-task competence. Our continual mechanism aims to preserve the actor's RL-acquired refinements while the actor and critic adapt to the current task.

\method retains the last $N$ training episodes of each completed task in a pooled memory $\mathcal{B}^{<k}$, giving $(k-1)N$ stored episodes at stage $k$. For a transition $j$ collected at stage $i<k$, $\refa_j$ is the action chunk generated by $\base^i$ at collection time and is retained unchanged. At each stage $k > 1$, each actor update uses a 1:1 ratio of current-task and prior-task samples, drawn in separate minibatches from $\mathcal{B}^k$ and $\mathcal{B}^{<k}$ for their respective actor losses. Before training stage $k$, we evaluate a frozen copy of the preceding-stage actor (the teacher $\pi_{\theta^{k-1}}$) on each retained transition to provide targets for the current actor (the student $\pi_{\theta^k}$). The actor may be deterministic or stochastic, depending on the off-policy learner. For distillation, we denote its deterministic evaluation output, obtained without action sampling or exploration noise, by $\bar{\pi}_\theta(s,\refa)$ and cache
\begin{equation}
  \hat{\mathbf{a}}^{\,k-1}_j = \bar{\pi}_{\theta^{k-1}}(s_j, \refa_j),
  \qquad j \in \mathcal{B}^{<k}.
  \label{eq:cached_targets}
\end{equation}
These targets remain fixed throughout stage $k$, and the teacher copy is discarded after caching. Behavioral targets are refreshed from the preceding-stage actor at each stage, while reference chunks retain their original values, keeping teacher and student at the same previously observed inputs $(s_j,\refa_j)$.

At stage $k$, TD updates on $\mathcal{B}^k$ adapt the inherited critic to $T^k$.
Although we could also replay $\mathcal{B}^{<k}$ for critic learning, only $N$ retained episodes per prior task may not reliably preserve its prior-task value estimates.
Applying the actor RL objective to these $\mathcal{B}^{<k}$ samples could therefore turn inaccurate critic estimates into drift in previously learned behavior. To avoid this risk, we apply critic TD and the actor RL objective only to $\mathcal{B}^k$, while using $\mathcal{B}^{<k}$ only to match the current actor to the preceding-stage actor at retained inputs,
\begin{align}
  \mathcal{L}_{\pi}^{\text{prior}}(\theta ; \mathcal{B}^{<k}) &=
    \mathop{\mathbb{E}}_{j \sim \mathcal{B}^{<k}}\Big[
    \big\lVert \pi_\theta(s_j, \refa_j)
    - \hat{\mathbf{a}}^{\,k-1}_j \big\rVert_2^2 \Big],
    \label{eq:actor_prior} \\
  \mathcal{L}_{\pi}(\theta;k) &= \mathcal{L}_{\pi}^{\text{cur}}(\theta;\mathcal{B}^{k})
    + \lambda\, \mathcal{L}_{\pi}^{\text{prior}}(\theta;\mathcal{B}^{<k}),
    \label{eq:actor_total}
\end{align}
where $\lambda$ weights prior-task distillation.
Thus, we do not use prior-task replay for critic updates or for the actor objective in Eq.~\eqref{eq:actor_current}.
Distillation constrains actor outputs at the stored $(s_j,\refa_j)$ inputs. Current-task interaction and deployment use fresh references from $\base^k$, which may differ even when the base policy retains earlier-task competence. We therefore evaluate prior-task retention after each stage with the updated base policy and actor together, testing whether the preserved behavior remains effective with current references.
\Cref{alg:main} summarizes one stage of \method.

\begin{algorithm}[!t]
\caption{One stage of \method}
\label{alg:main}
\begin{algorithmic}[1]
  \REQUIRE base policy $\base^{k}$, frozen visual encoder $E$, chunk length $H$, handoff $(\theta^{k-1}, \psi^{k-1}, \mathcal{B}^{<k})$
  \STATE Initialize $\pi_\theta, Q_\psi$ from $\theta^{k-1}, \psi^{k-1}$
  \STATE Cache $\hat{\mathbf{a}}^{\,k-1}_j = \bar{\pi}_{\theta^{k-1}}(s_j, \refa_j)$
         for $j \in \mathcal{B}^{<k}$~\eqcomment{eq:cached_targets}
  \STATE Collect warmup rollouts from $\base^{k}$ 
  \FOR{$t = 0, H, 2H, \dots$}
    \STATE $s_t \leftarrow (E(\mathbf{I}_t), \mathbf{q}_t)$,\ \
           $\refa_t \sim \base^{k}(\cdot \mid \mathbf{I}_t, \mathbf{q}_t, \ell^k)$
    \STATE Execute $\mathbf{a}_t \sim \pi_\theta(\cdot \mid s_t, \refa_t)$ and observe $R_t^{(H)}$, $s_{t+H}$
    \STATE Store $(s_t, \mathbf{a_t}, \refa_t, R_t^{(H)}, s_{t+H}, \refa_{t+H})$ in $\mathcal{B}^k$
    \STATE Update $\psi$ on $\mathcal{B}^k$~\eqcomment{eq:td_loss}
    \STATE Update $\theta$ on $\mathcal{B}^k \cup \mathcal{B}^{<k}$~\eqcomment{eq:actor_total}
  \ENDFOR
  \RETURN last $N$ episodes from $\mathcal{B}^k$,  handoff $(\theta^{k}, \psi^{k})$
\end{algorithmic}
\end{algorithm}

\begin{figure}[!t]
  \centering
  \begin{subfigure}[t]{0.33\textwidth}
    \centering
    \includegraphics[width=\linewidth]{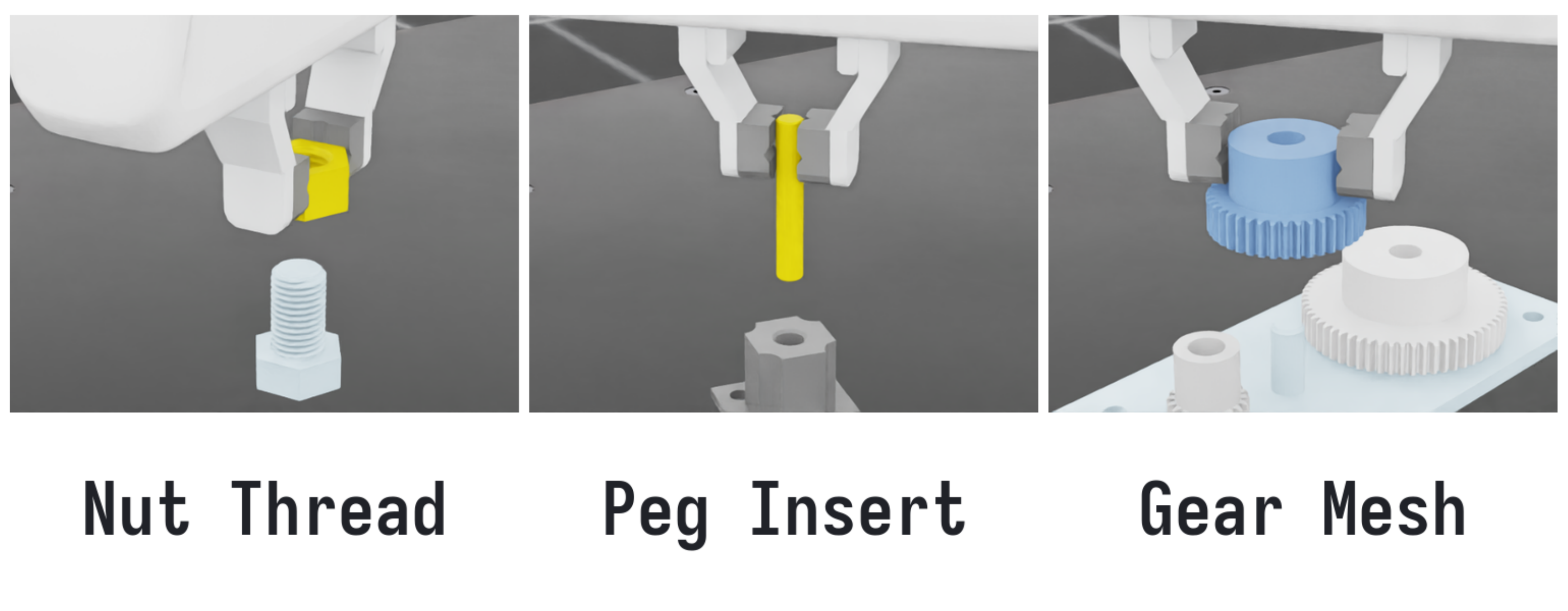}
    \caption{FORGE \cite{forge}}

    \label{fig:tasks_forge}
  \end{subfigure}\hfill%
  \begin{subfigure}[t]{0.65\textwidth}
    \centering
    \includegraphics[width=\linewidth]{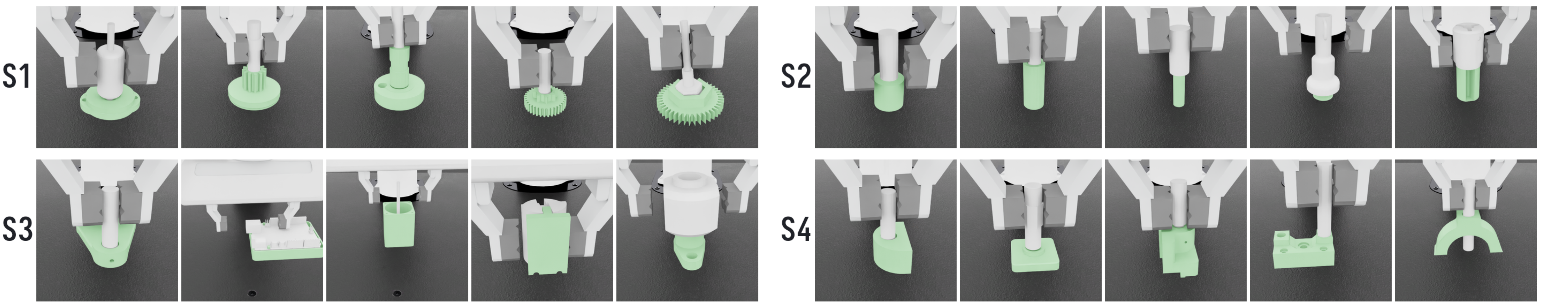}

    \caption{AutoMate \cite{automate}}
    \label{fig:tasks_automate}
  \end{subfigure}

  \par\vspace{6pt}
  \begin{subfigure}[t]{0.50\textwidth}
    \centering
    \includegraphics[width=\linewidth]{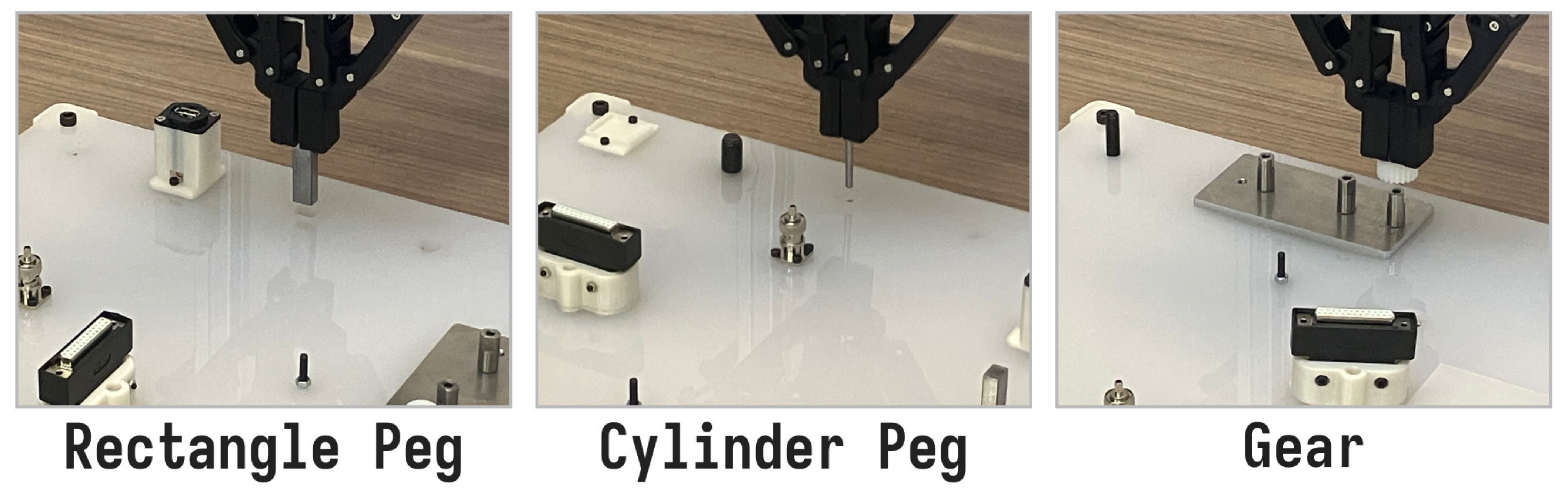}

    \caption{Real-world NIST Assembly Task Board}
    \label{fig:tasks_nist}
  \end{subfigure}\hfill%
  \begin{subfigure}[t]{0.49\textwidth}
    \centering
    \includegraphics[width=\linewidth]{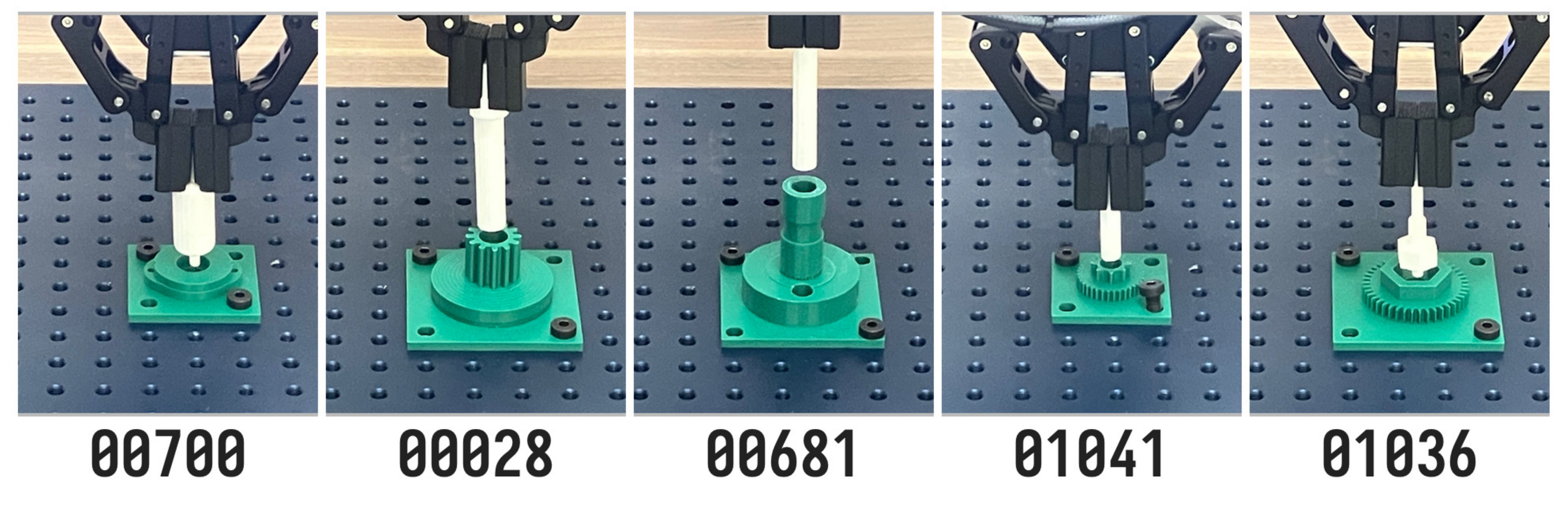}

    \caption{Real-world AutoMate}
    \label{fig:tasks_automate_real}
  \end{subfigure}

  \caption{%
    \textbf{Overview of experimental tasks.}
    (a) Three contact-rich manipulation tasks in FORGE \cite{forge}.
    Four orderings are evaluated:
    S1: \textbf{Nut Thread $\rightarrow$ Peg Insert $\rightarrow$ Gear Mesh},
    S2: \textbf{Gear Mesh $\rightarrow$ Peg Insert $\rightarrow$ Nut Thread},
    S3: \textbf{Peg Insert $\rightarrow$ Gear Mesh $\rightarrow$ Nut Thread}, and
    S4: \textbf{Peg Insert $\rightarrow$ Nut Thread $\rightarrow$ Gear Mesh}.
    (b) Four five-stage AutoMate sequences in simulation:
    S1: \textbf{00700 $\rightarrow$ 00028 $\rightarrow$ 00681 $\rightarrow$ 01041 $\rightarrow$ 01036},
    S2: \textbf{01092 $\rightarrow$ 00141 $\rightarrow$ 00255 $\rightarrow$ 00486 $\rightarrow$ 00615},
    S3: \textbf{00731 $\rightarrow$ 00470 $\rightarrow$ 00117 $\rightarrow$ 00726 $\rightarrow$ 00296}, and
    S4: \textbf{00417 $\rightarrow$ 00083 $\rightarrow$ 01079 $\rightarrow$ 00514 $\rightarrow$ 01029}.
    (c) Three insertion task sequence on the NIST Assembly Task Board and
    (d) Five-stage AutoMate sequence executed on real 3D-printed assets.
  }
  \label{fig:experimental-tasks}
\end{figure}

\section{Experiments}
\label{sec:exp}

Our experiments ask three questions. 1) Does \method retain earlier assembly skills once the base policy has been updated for a later stage? 2) Does inheriting the lightweight actor and critic accelerate a new stage? And 3) Which of \method's components is responsible for each effect? Sec.~\ref{subsec:setup} describes what is held common across platforms. Sec.~\ref{subsec:sim} and Sec.~\ref{subsec:real} address the first two questions in simulation and on real-world DROID~\cite{khazatsky2024droid}. The two settings use different base policies, reinforcement learning algorithms, and interaction budgets, which lets us verify that the findings do not depend on a particular setup. Sec.~\ref{subsec:ablation} addresses the third question in simulation, where the stage budget and the task sequence can be held fixed across variants.

\subsection{Experimental Setup}
\label{subsec:setup}


\paragraph*{Baselines} We compare continual learners within a shared off-policy actor--critic architecture, alongside reference methods with different learning or storage requirements. In the main simulation comparisons, all RL methods use the same GR00T N1.7 backbone and per-stage environment-step budget. The matched shared actor--critic methods additionally share the cumulative-demonstration fine-tuning protocol of Sec.~\ref{subsec:problem} and the off-policy RL algorithm, and inherit the actor--critic pair across tasks. \textbf{Sequential} trains it using only current-task data. \textbf{ER}~\cite{chaudhry2019tiny} retains the same number of episodes as \method per task, but applies the RL objective to the retained memory rather than using actor distillation. \textbf{EWC}~\cite{kirkpatrick2017overcoming} penalizes actor drift from the previous-stage parameters, while \textbf{RETAIN}~\cite{yadav2026robust} merges the current and previous actor parameters through discounted interpolation. As an external reference, \textbf{Simple Recipe}~\cite{hu2026vlacrl} sequentially fine-tunes the VLA directly with on-policy RL. \textbf{Single-task RL} initializes and trains a new actor--critic pair for each task. Finally, \textbf{Isolated} initializes each new pair from the preceding stage but retains a separate checkpoint per task, avoiding shared-parameter interference at the cost of storing one learner per task.

\paragraph*{Metrics} Following prior work~\cite{liu2023libero, hu2026vlacrl, liu2026forgetmenot}, forward transfer (FWT) averages training success on new tasks, and gains over Single-task RL indicate transfer between tasks. Negative backward transfer (NBT) measures the performance on preceding tasks from the current stage, where lower values indicate better backward retention. Continual AUC combines learning efficiency and retention. Final accuracy and forgetting describe end-of-sequence success and prior-task loss.

Stages share an environment-step budget and uniform evaluation grid $\mathcal{E}$. Let $c_{i,j,e}$ be task $j$'s success rate after $e$ steps in stage $i$. Set $c_{i,i}=\max_{e}c_{i,i,e}$ and $c_{i,j}=c_{i,j,e^{*}_{i}}$, using the earliest maximizing checkpoint $e^{*}_{i}$.
\begin{equation*}
\begin{aligned}
\mathrm{FWT}_{k} &= \tfrac{1}{|\mathcal{E}|}\textstyle\sum_{e\in\mathcal{E}} c_{k,k,e}, \\
\mathrm{NBT}_{k} &= \tfrac{1}{K-k}\textstyle\sum_{m=k+1}^{K}\big(c_{k,k}-c_{m,k}\big), \\
\mathrm{AUC}_{k} &= \tfrac{1}{K-k+1}\big(\mathrm{FWT}_{k}+\textstyle\sum_{m=k+1}^{K} c_{m,k}\big).
\end{aligned}
\end{equation*}
We average FWT and continual AUC over ${k\in[K]}$, and NBT over $k\in[K-1]$. Final deployment metrics are $\mathrm{ACC}_{\mathrm{final}}=\tfrac{1}{K}\sum_{k}c_{K,k}$ and $\smash{\mathrm{F}_{\mathrm{final}}=\tfrac{1}{K-1}\sum_{k<K}(c_{k,k}-c_{K,k})}$. Unlike NBT, $\smash{\mathrm{F}_{\mathrm{final}}}$ ignores transient performance drops and measures forgetting only at the end of the sequence. These two metrics describe the end-of-sequence deployment performance.

\paragraph*{Hyperparameter selection in \method} We choose $\lambda$ in Eq.~\ref{eq:actor_total} to balance prior-task preservation against current-task adaptation, favoring stronger distillation when successive tasks share fewer behaviors. Accordingly, we use $\lambda=1.0$ for the more diverse real-world NIST sequence and $\lambda=0.1$ for the more closely related real-world AutoMate assemblies. The coefficient $\beta$ in Eq.~\ref{eq:actor_current} controls the penalty on deviation from the base policy's action proposals. We use $\beta=0.1$ in simulation and increase it to $10.0$ on hardware, where keeping refinements close to the VLA's proposals is particularly important during physical contact.

\newcommand{\curvewithtask}[4][0.065\linewidth]{%
  \begin{tikzpicture}
    \node[inner sep=0, outer sep=0] (curve) {%
      \includegraphics[width=\linewidth]{#4}%
    };
    \node[
      anchor=north west,
      xshift=#1,
      yshift=#2,
      inner sep=0.7pt,
      outer sep=0,
      fill=white,
      draw=black!25,
      line width=0.2pt
    ] at (curve.north west) {%
      \includegraphics[
        width=0.20\linewidth,
        height=0.16\linewidth,
        keepaspectratio
      ]{#3}%
    };
  \end{tikzpicture}%
}

\subsection{Simulation Experiments}
\label{subsec:sim}

\paragraph*{Benchmarks} We use AutoMate~\cite{automate} and FORGE~\cite{forge}, both built on Isaac Lab~\cite{mittal2025isaac}. AutoMate provides assembly tasks in which plugs of different
shapes must be inserted into matching sockets. We organize these tasks into four five-stage sequences. FORGE provides three contact-rich tasks (peg insertion, gear meshing, and nut threading), which we run as a three-stage sequence. For both tasks, the policy operates in a 6D end-effector delta action space. As task ordering is known to affect continual learning outcomes~\cite{liu2023libero}, we report the aggregated results over multiple sequences. Specifically, we aggregate over all four asset sequences for AutoMate, and we sample four task orderings randomly for FORGE; every result is aggregated over 4 seeds. \Cref{fig:tasks_forge,fig:tasks_automate} visualizes the tasks.

\paragraph*{Setup} Each stage runs 64 parallel environments for a budget of 500K environment steps, is evaluated every 50K steps, and terminates an episode on success. The reward is sparse and emitted only on task success. Since an episode ends at the success step, we label the final three steps of a successful episode with a reward of 1 so that the terminal chunk carries a learning signal, and we use a discount factor of $\gamma = 0.997$. We use GR00T N1.7~\cite{nvidia2025groot} for $\pi_{\text{base}}$, which is fully fine-tuned at each stage on 1 demonstration for 1000 gradient steps on AutoMate, and on 10, 20, and 50 demonstrations for gear meshing, nut threading, and peg insertion, respectively, for 2000 steps on FORGE. These task-specific demonstration budgets aim to provide a reasonably competent base policy while minimizing human demonstration collection, reserving further improvement for online interactions in RL. We further examine the effect of fewer task demonstrations in \Cref{subsec:ablation}. For RL, we adopt FastSAC~\cite{seo2025fasttd3} over chunks of length $H=16$ on frozen DINOv3 features~\cite{simeoni2026dinov}, with 50 (AutoMate) / 100 (FORGE) retained episodes per stage.

\begin{table}[!htb]
    \centering
    \small
    \caption{Results on FORGE, aggregated over 4 task orderings and 4 seeds. Bold and underlined values denote the best and second-best results among the matched shared actor--critic methods. 
    }
    \label{tab:forge}
    \setlength{\tabcolsep}{4pt}
    \begin{tabular*}{\textwidth}{@{\extracolsep{\fill}}lccccc@{}}
    \toprule
    Method & NBT $\downarrow$ & FWT $\uparrow$ & AUC $\uparrow$ & ACC$_{\text{final}}$ $\uparrow$ & F$_{\text{final}}$ $\downarrow$ \\
    \midrule
    \multicolumn{6}{@{}l}{\emph{Shared actor--critic methods}} \\
    Sequential & 0.427 & 0.819 & \underline{0.718} & \underline{0.685} & 0.409 \\
    EWC~\cite{kirkpatrick2017overcoming} & \underline{0.217} & 0.593 & 0.582 & 0.603 & \underline{0.224} \\
    RETAIN~\cite{yadav2026robust} & 0.514 & 0.802 & 0.682 & 0.644 & 0.474 \\
    ER~\cite{chaudhry2019tiny} & 0.518 & \textbf{0.833} & 0.697 & 0.626 & 0.516 \\
    \textbf{\method (Ours)} & \textbf{0.065} & \underline{0.823} & \textbf{0.869} & \textbf{0.917} & \textbf{0.068} \\
    \midrule
    \multicolumn{6}{@{}l}{\emph{Reference methods}} \\
    $\pi_{\text{base}}$ & -- & -- & -- & 0.687 & -- \\
    Single-task RL & -- & 0.741 & -- & -- & -- \\
    Simple Recipe~\cite{hu2026vlacrl} & 0.038 & 0.586 & 0.583 & 0.594 & 0.061 \\
    Isolated$^\dagger$ & -- & 0.816 & 0.868 & 0.919 & -- \\
    \bottomrule
    \end{tabular*}
\end{table}

\begin{table}[!htb]
    \centering
    \small
    \caption{Results on AutoMate, aggregated over 4 asset sequences and 4 seeds. Bold and underlined values denote the best and second-best results among the matched shared actor--critic methods. 
    }
    \label{tab:automate}
    \setlength{\tabcolsep}{4pt}
    \begin{tabular*}{\textwidth}{@{\extracolsep{\fill}}lccccc@{}}
    \toprule
    Method
    & NBT $\downarrow$
    & FWT $\uparrow$
    & AUC $\uparrow$
    & ACC$_{\text{final}}$ $\uparrow$
    & F$_{\text{final}}$ $\downarrow$ \\
    \midrule
    \multicolumn{6}{l}{\emph{Shared actor--critic methods}} \\
    Sequential & 0.178 & 0.824 & 0.792 & 0.765 & 0.175 \\
    EWC~\cite{kirkpatrick2017overcoming}        & \underline{0.075} & 0.704 & 0.704 & 0.699 & \underline{0.067} \\
    RETAIN~\cite{yadav2026robust}     & 0.191 & 0.822 & 0.784 & 0.750 & 0.192 \\
    ER~\cite{chaudhry2019tiny}         & 0.167 & \underline{0.834} & \underline{0.799} & \underline{0.779} & 0.176 \\
    \textbf{\method (Ours)}  & \textbf{0.047} & \textbf{0.848} & \textbf{0.858} & \textbf{0.863} & \textbf{0.054} \\
    \midrule
    \multicolumn{6}{l}{\emph{Reference methods}} \\
    $\pi_{\text{base}}$ & -- & -- & -- & 0.757 & -- \\
    Single-task RL & -- & 0.807 & -- & -- & -- \\
    Simple Recipe~\cite{hu2026vlacrl} & 0.016 & 0.730 & 0.726 & 0.730 & 0.019 \\
    Isolated$^\dagger$ & -- & 0.826 & 0.858 & 0.879 & -- \\
    \bottomrule
    \end{tabular*}%
\end{table}

\paragraph*{Results} Among the matched shared actor--critic methods, \method achieves the lowest NBT and final forgetting and the highest AUC and final accuracy on both benchmarks (Tables~\ref{tab:forge} and~\ref{tab:automate}). It also exceeds Single-task RL in FWT under the same stage-wise interaction budget. EWC trades new-task learning for retention, whereas Sequential, ER, and RETAIN learn new tasks efficiently but forget more. On AutoMate, \method approaches the privileged Isolated reference while attaining higher FWT and maintaining a single shared actor--critic pair. As an external reference, Simple Recipe attains lower FWT and continual AUC than \method on both benchmarks under the same VLA backbone and interaction budget (Tables~\ref{tab:forge} and~\ref{tab:automate}). These results indicate the efficacy of \method in setup with limited interaction budget.



\begin{figure}[!t]
  \centering

  \makebox[\textwidth][l]{%
    \begin{minipage}[c]{0.18\textwidth}
      \includegraphics[width=\linewidth]{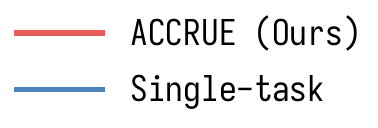}
    \end{minipage}%
    \hspace{0.015\textwidth}%
    \begin{subfigure}[c]{0.188\textwidth}
      \centering
      \curvewithtask[0.29\linewidth]
        {-0.02\linewidth}
        {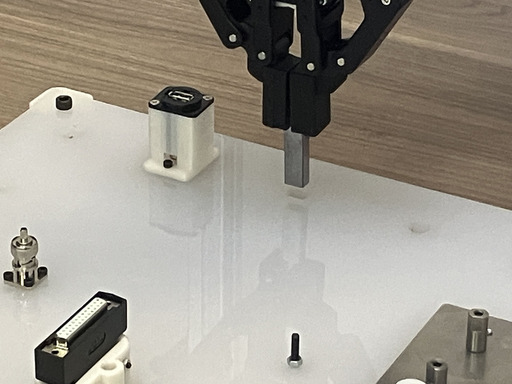}
        {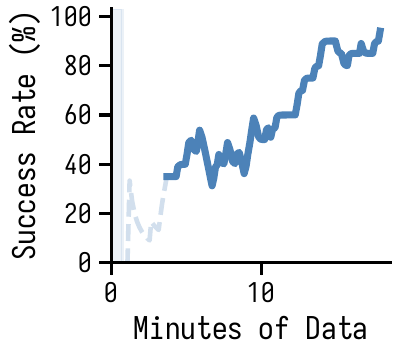}
    \end{subfigure}%
    \hspace{0.025\textwidth}%
    \begin{subfigure}[c]{0.188\textwidth}
      \centering
      \curvewithtask
        {-0.02\linewidth}
        {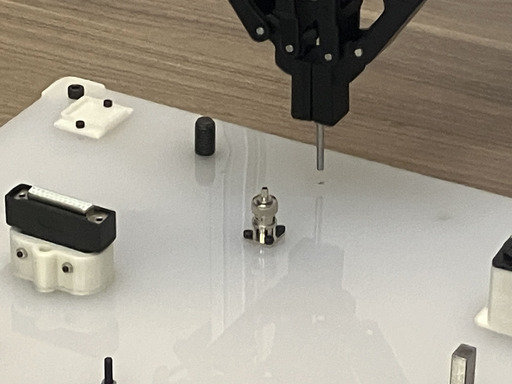}
        {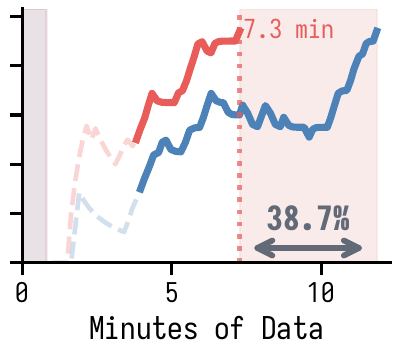}
    \end{subfigure}%
    \hspace{0.025\textwidth}%
    \begin{subfigure}[c]{0.188\textwidth}
      \centering
      \curvewithtask
        {-0.02\linewidth}
        {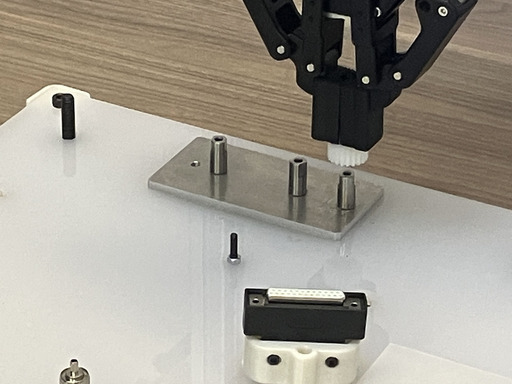}
        {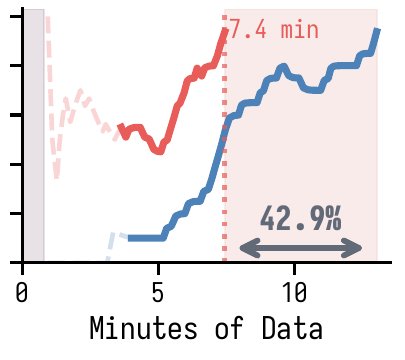}
    \end{subfigure}%
  }

  \par

  \begin{subfigure}[t]{0.188\textwidth}
    \centering
    \curvewithtask[0.29\linewidth]
      {-0.02\linewidth}
      {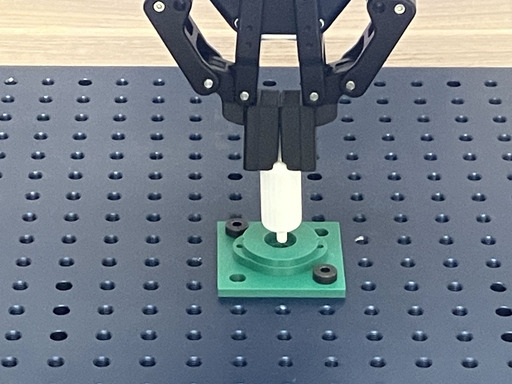}
      {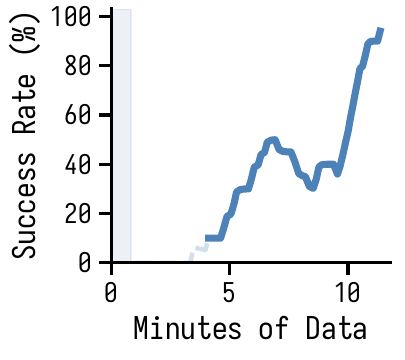}
  \end{subfigure}
  \hfill
  \begin{subfigure}[t]{0.188\textwidth}
    \centering
    \curvewithtask
      {-0.02\linewidth}
      {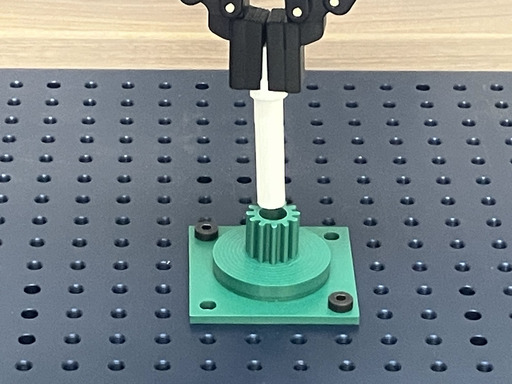}
      {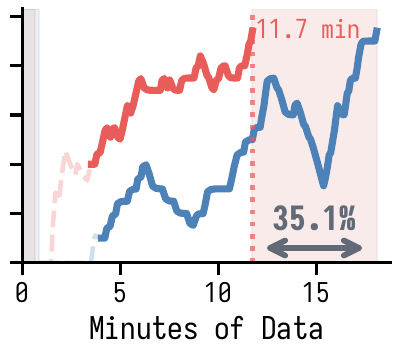}
  \end{subfigure}
  \hfill
  \begin{subfigure}[t]{0.188\textwidth}
    \centering
    \curvewithtask
      {-0.48\linewidth}
      {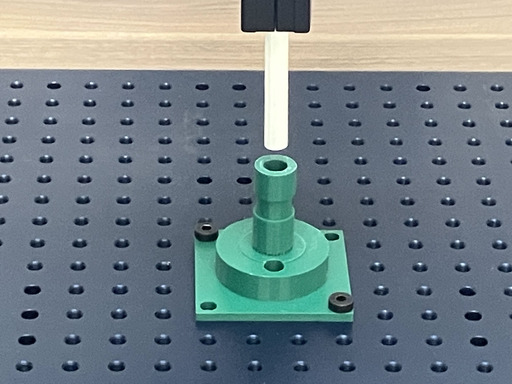}
      {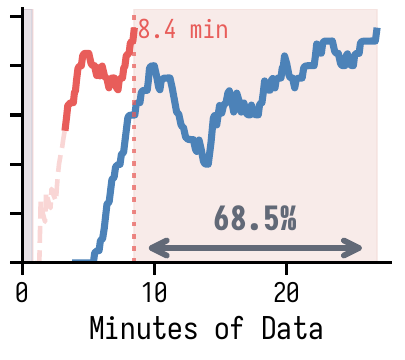}
  \end{subfigure}
  \hfill
  \begin{subfigure}[t]{0.188\textwidth}
    \centering
    \curvewithtask
      {-0.48\linewidth}
      {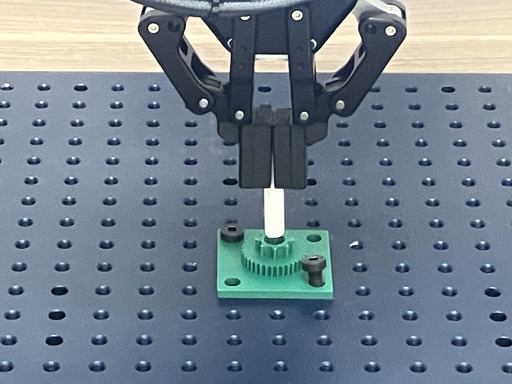}
      {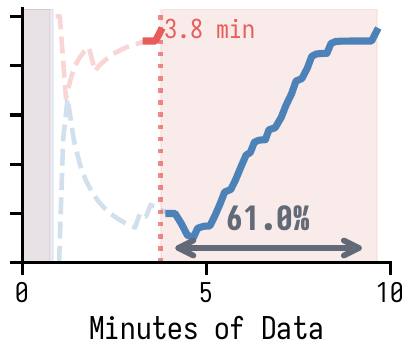}
  \end{subfigure}
  \hfill
  \begin{subfigure}[t]{0.188\textwidth}
    \centering
    \curvewithtask
      {-0.02\linewidth}
      {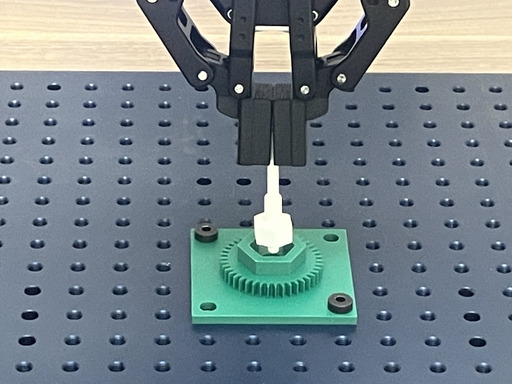}
      {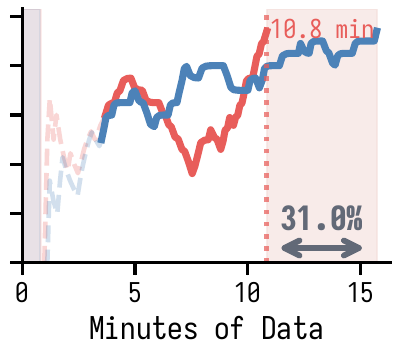}
  \end{subfigure}

  \caption{%
    \textbf{Real-world continual RL on DROID setup.}
    (Top) Three-stage NIST Assembly Task Board sequence:
    \textbf{Rectangle Peg $\rightarrow$ Cylinder Peg $\rightarrow$ Gear}.
    (Bottom) Five-stage AutoMate sequence:
    \textbf{00700 $\rightarrow$ 00028 $\rightarrow$ 00681 $\rightarrow$
    01041 $\rightarrow$ 01036}.
    Each panel reports the running success rate over a 20-episode
    window as a function of cumulative robot interaction time for RL,
    excluding human demonstration collection.
    Single-task trains the actor and critic from scratch, whereas
    \method inherits them from the previous stage.
    Gray shading marks warmup rollouts collected by the base policy
    to initialize the replay buffer, and dashed segments at the left indicate that fewer than 20 episodes have been collected.
    The dotted line marks when Continual reaches the stopping
    threshold of 95\% success (19/20 episodes), and the shaded span
    reports robot interaction time saved relative to Single-task
    reaching the same threshold.
    The first stage of each sequence is shared by both conditions.
  }
  \label{fig:real_curves}
\end{figure}

\subsection{DROID Experiments}
\label{subsec:real}

\paragraph*{Tasks} To validate \method beyond simulation, We assess it on the DROID platform~\cite{khazatsky2024droid} with two task families. The first sequences three insertions on the NIST Assembly Task Board (rectangular peg, cylindrical peg, and gear mesh). The second runs a five-stage sequence of AutoMate assemblies on the same hardware, which tests whether the simulation task family transfers to a physical setup. For both tasks, the policy operates in a 6D end-effector delta action space, and each episode begins with the target object held by an object-specific 3D-printed gripper. \Cref{fig:tasks_automate_real,fig:tasks_nist} visualizes the tasks.

\paragraph*{Setup} For the base policy $\pi_{\text{base}}$, we utilize $\pi_{0.5}$~\cite{black2025pi05} rather than GR00T N1.7 owing to its stronger performance on DROID. Before RL at each stage, we fine-tuned the model with LoRA on 5 (AutoMate) and 10 (NIST) demonstrations collected by SpaceMouse teleoperation per each task. Unlike in simulation, the low-dimensional input contains only the robot's seven joint positions, without additional object- or task-state information. For RL backbone, we employ FastTD3~\cite{seo2025fasttd3} rather than FastSAC, because its explicit control of exploration noise avoids erratic contact behavior of the arm. We use actor and critic over chunks of length $H=10$, with 20 retained episodes per stage. The reward is sparse on episode success, labeled by an operator. An asynchronous learner and actor process, similar to~\cite{dong2026expoft}, keeps data collection running while the learner is busy. We report a running success rate over a window of 20 episodes and terminate a stage once it reaches 95\% (19/20). Retention is measured by 20 evaluation trials on each earlier task at the end of each later stage. We compare against single-task RL within the same framework, since a full baseline sweep is not feasible at this scale.

\begin{table}[!htbp]
    \centering
    \caption{Real-world retention reporting successes over 20 evaluation trials on each earlier task at the end of each later stage. 
    }
    \label{tab:real_retention}
    \small
    \setlength{\tabcolsep}{5pt}
    \begin{tabular}{lcccc}
    \toprule
     & \multicolumn{4}{c}{After stage} \\
    \cmidrule(lr){2-5}
     & 2 & 3 & 4 & 5 \\
    \midrule
    \multicolumn{5}{l}{\textit{NIST board}} \\
    Stage 1 (Rectangle Peg) & 20 & 19 & & \\
    Stage 2 (Cylinder Peg) & -- & 20 & & \\
    \midrule
    \multicolumn{5}{l}{\textit{AutoMate}} \\
    Stage 1 (00700) & 20 & 19 & 18 & 19 \\
    Stage 2 (00028) & -- & 19 & 19 & 20 \\
    Stage 3 (00681) & -- & -- & 19 & 20 \\
    Stage 4 (01041) & -- & -- & -- & 20 \\
    \bottomrule
    \end{tabular}
\end{table}

\paragraph*{Results} Inheriting the actor and critic reduces the robot time needed to reach the stopping criterion by 31.0 to 68.5 percent across the stages of the two sequences (Fig.~\ref{fig:real_curves}), and retention remains essentially complete, with every earlier task succeeding on at least 18 of 20 trials and 19.4 on average (Table~\ref{tab:real_retention}). Both effects are larger than in simulation, which we attribute to the far smaller interaction budget: a better starting point is worth more when a stage is measured in tens of minutes of robot time, and the saving is the difference between a stage that fits in a single session and one that does not. Notably, these results are obtained with a different base policy family and a different RL algorithm than in simulation, under real sensing and contact noise, indicating that \method depends on neither a particular base policy nor a particular RL algorithm.






\subsection{Ablation Studies and Analyses}
\label{subsec:ablation}


\begin{table}[!htbp]
    \centering
    \small
    \caption{Ablation of the design choices on FORGE sequence S1, holding the task sequence, stage budget, and all shared hyperparameters fixed.
    Best values are in bold and second best are underlined.
    }
    \label{tab:ablation}
    \setlength{\tabcolsep}{4pt}
    \begin{tabular*}{\textwidth}{@{\extracolsep{\fill}}lccccc@{}}
    \toprule
    Method & NBT $\downarrow$ & FWT $\uparrow$ & AUC $\uparrow$ & ACC$_{\text{final}}$ $\uparrow$ & F$_{\text{final}}$ $\downarrow$ \\
    \midrule
    \method & \textbf{0.028} & \textbf{0.854} & \textbf{0.906} & \textbf{0.961} & \textbf{0.037} \\
    DINOv3 $\rightarrow$ GR00T VLM & 0.071 & 0.814 & 0.846 & 0.883 & 0.113 \\
    $\pi_{\theta}(s, \refa) \rightarrow \pi_{\theta}(s)$ & \underline{0.035} & 0.807 & 0.869 & 0.936 & 0.050 \\
    $\beta = 0.0$ & \textbf{0.029} & 0.763 & 0.821 & 0.942 & \textbf{0.037} \\
    Reset Critic & 0.042 & 0.841 & \underline{0.890} & \underline{0.942} & \underline{0.054} \\
    Update Critic w/ $\mathcal{B}^{<k}\cup\mathcal{B}^k$ & 0.037 & \underline{0.842} & 0.888 & 0.925 & 0.067 \\
    \bottomrule
    \end{tabular*}
\end{table}

\paragraph*{Actor--critic inputs} Replacing the fixed DINOv3 representation with features extracted by the stage-updated GR00T VLM yields worse retention and forward transfer (Table~\ref{tab:ablation}). Although this comparison also changes the encoder itself and therefore does not isolate representation drift, it supports the practical advantage of a fixed visual interface for inheriting the actor and critic across base-policy updates. Removing the reference-action input likewise lowers retention and transfer, proving that reference action provides useful conditioning beyond the visual state alone.

\paragraph*{Reference guidance and critic inheritance} Removing the reference-deviation penalty by setting $\beta=0$ in Eq.~\ref{eq:actor_current} has little effect on backward retention but substantially weakens forward transfer and task learning (Table~\ref{tab:ablation}). Explicitly biasing the actor toward the reference action therefore supports efficient local refinement rather than preservation of prior behavior. Resetting the critic at each stage produces modestly more forgetting and lower task-learning efficiency. Extending critic updates to prior-task replay increases forgetting and lowers continual AUC and final success, supporting our choice to train the critic only on current-task transitions.

\begin{table}[!htbp]
    \centering
    \caption{Effect of the number of retained episodes per stage $N$ in FORGE sequence S1. Best values are in bold and second best are underlined.}
    \label{tab:replay}
    \small
    \setlength{\tabcolsep}{4pt}
    \begin{tabular*}{\textwidth}{@{\extracolsep{\fill}}lccccc@{}}
    \toprule
    $N$ & NBT $\downarrow$ & FWT $\uparrow$ & AUC $\uparrow$ & ACC$_{\text{final}}$ $\uparrow$ & F$_{\text{final}}$ $\downarrow$ \\
    \midrule
    10  & 0.107 & \underline{0.853} & 0.854 & 0.847 & 0.192 \\
    20 & 0.093 & 0.844 & 0.859 & 0.869 & 0.167 \\
    50 & 0.060 & 0.838 & 0.871 & 0.900 & 0.104 \\
    100 & \textbf{0.040} & \textbf{0.854} & \textbf{0.896} & \textbf{0.925} & \textbf{0.062} \\
    200 & \underline{0.043} & 0.842 & \underline{0.879} & \underline{0.903} & \underline{0.083} \\
    \bottomrule
    \end{tabular*}
\end{table}

\paragraph*{Replay budget} Retaining $N=100$--$200$ episodes per completed task (approximately 3\% of the total replay buffer) provides the strongest joint retention--learning trade-off (Table~\ref{tab:replay}). Memories of $10$--$20$ episodes are insufficient to preserve prior behavior, while increasing the budget from $100$ to $200$ episodes degrades performance. This non-monotonic trend shows that effective continual post-training does not follow directly from replaying more prior data.

\begin{table}[!htbp]
      \centering
    \small
      \caption{Ablation of demo retention during sequential fine-tuning of $\pi_{\text{base}}$ on FORGE sequence S1. Best values are shown in bold and second-best values are underlined.}
      \label{tab:demo-retention}
      \setlength{\tabcolsep}{4pt}
      \begin{tabular*}{\textwidth}{@{\extracolsep{\fill}}rccccccc@{}}
      \toprule
      & \multicolumn{2}{c}{$\pi_{\text{base}}$ (GR00T N1.7)}
      & \multicolumn{5}{c}{\method} \\
      \cmidrule(lr){2-3}
      \cmidrule(lr){4-8}
      Demo Retention
      & SR $\uparrow$
      & NBT $\downarrow$
      & NBT $\downarrow$
      & FWT $\uparrow$
      & AUC $\uparrow$
      & ACC$_{\text{final}}$ $\uparrow$
      & F$_{\text{final}}$ $\downarrow$ \\
      \midrule
      100\%
      & \textbf{0.710}
      & \textbf{0.013}
      & \textbf{0.028}
      & \textbf{0.854}
      & \textbf{0.906}
      & \textbf{0.961}
      & \textbf{0.037} \\
      50\%
      & 0.580
      & \underline{0.035}
      & \underline{0.056}
      & 0.817
      & \underline{0.860}
      & \underline{0.908}
      & \underline{0.092} \\
      10\%
      & \underline{0.637}
      & 0.090
      & 0.078
      & 0.811
      & 0.844
      & 0.875
      & 0.133 \\
      5\%
      & 0.577
      & 0.163
      & 0.071
      & \underline{0.836}
      & 0.859
      & 0.894
      & 0.113 \\
      \bottomrule
      \end{tabular*}
  \end{table}

\paragraph*{Robustness to base-policy forgetting} We vary the fraction of prior-task demonstrations retained for sequential VLA fine-tuning from 100\% to 5\%, keeping the continual RL procedure and its replay budget fixed (Table~\ref{tab:demo-retention}). Reduced demonstration replay degrades both base-policy and post-RL performance relative to using all demonstrations. Nevertheless, \method mitigates forgetting and maintains new-task learning efficiency even when the base policy itself forgets.

\begin{table}[!htbp]
    \centering
    \small
    \caption{\method with different $\pi_{\text{base}}$ in FORGE sequence S1.}
    \label{tab:wam}
    \setlength{\tabcolsep}{4pt}
    \begin{tabular*}{\textwidth}{@{\extracolsep{\fill}}lcccccc@{}}
    \toprule
    Method & Base SR & NBT $\downarrow$ & FWT $\uparrow$ & AUC $\uparrow$ & ACC$_{\text{final}}$ $\uparrow$ & F$_{\text{final}}$ $\downarrow$ \\
    \midrule
    \multicolumn{7}{l}{\emph{\textcolor{gray}{$\pi_{\text{base}}$: GR00T N1.7~\cite{nvidia2025groot} (VLA)}}}\\
    ER & \multirow{2}{*}{0.710}  & 0.339 & 0.840 & 0.705 & 0.611 & 0.529 \\
    \method & & \textbf{0.028} & \textbf{0.854} & \textbf{0.906} & \textbf{0.961} & \textbf{0.037} \\
    \midrule
    \multicolumn{7}{l}{\emph{\textcolor{gray}{$\pi_{\text{base}}$: Fast-WAM~\cite{yuan2026fast} (WAM)}}}\\
    ER & \multirow{2}{*}{0.347}  & 0.419 & 0.742 & 0.588 & 0.517 & 0.625 \\
    \method &  & \textbf{0.035} & \textbf{0.768} & \textbf{0.847} & \textbf{0.905} & \textbf{0.042} \\
    \bottomrule
    \end{tabular*}
\end{table}

\paragraph*{Base-policy dependence} In Table~\ref{tab:wam}, we apply \method on Fast-WAM~\cite{yuan2026fast} to prove the effectiveness regardless of model architecture. \method maintains strong retention and final performance even with Fast-WAM despite their different architecture and pre-RL success rates. ER similarly degrades under both policy families.

\section{Conclusion}
\label{sec:conclusion}

We presented \method, a continual off-policy RL framework for robot foundation models. Our approach is to inherit lightweight actor--critic guided by RFM's reference actions and to apply prior-task distillation. Through this, \method enables efficient RL on new tasks while preserving earlier behavior with limited replay. Experiments in simulation and on real robots demonstrate faster subsequent-task learning with significantly reduced forgetting.
These results show that continual RL can support more sample-efficient post-training of robot foundation models as new tasks arrive, while retaining previously learned behaviors.

Our work establishes advances on task sequences of up to five stages in contact-rich assembly, with replay memory bounded per task.
We leave it for future work to scale \method to longer task sequences under a fixed total memory budget. Additionally, a natural next step is to evaluate \method on a broader range of manipulation tasks, including less closely related tasks. Finally, it is promising to learn shared action representation and transfer RL post-training across varied embodiments or action spaces.
Overall, \method shows that RL-acquired knowledge can accumulate across sequential deployment stages, providing a practical path to continually learn skills on robotic foundation models.












\bibliographystyle{plainnat}
\bibliography{references}

\end{document}